# Network Fragments:
# Representing Knowledge for Constructing Probabilistic Models

**Kathryn Blackmond Laskey**
Department of Systems Engineering and C³I Center
George Mason University
Fairfax, VA 22030
klaskey@gmu.edu

**Suzanne M. Mahoney**
Information Extraction and Transport, Inc.
1730 N. Lynn Street, Suite 502
Arlington, VA 22209
suzanne@iet.com

## Abstract

In most current applications of belief networks, domain knowledge is represented by a single belief network that applies to all problem instances in the domain. In more complex domains, problem-specific models must be constructed from a knowledge base encoding probabilistic relationships in the domain. Most work in knowledge-based model construction takes the rule as the basic unit of knowledge. We present a knowledge representation framework that permits the knowledge base designer to specify knowledge in larger semantically meaningful units which we call network fragments. Our framework provides for representation of asymmetric independence and canonical intercausal interaction. We discuss the combination of network fragments to form problem-specific models to reason about particular problem instances. The framework is illustrated using examples from the domain of military situation awareness.

## 1 INTRODUCTION

The vast majority of published applications of belief networks consist of *template models*. A template model is appropriate for problem domains in which the relevant variables, their state spaces, and their probabilistic relationships do not vary from problem instance to problem instance. Thus, generic knowledge about the domain can be represented by a fixed belief network over a fixed set of variables, obtained by some combination of expert judgment and learning from observation. Problem solving for a particular case is performed by conditioning the network on case-specific evidence and computing the posterior distributions of variables of interest. For example, a medical diagnosis template network would contain variables representing background information about a patient, possible medical conditions the patient might be experiencing, and clinical findings that might be observed. The network encodes probabilistic relationships among these variables. To perform diagnosis on a particular patient, background information and findings for the patient are entered as evidence and the posterior probabilities of the possible medical conditions are reported. Although values of the evidence variables vary from patient to patient, the relevant variables and their probabilistic relationships are assumed to be the same for all patients. It is this assumption that justifies the use of template models.

The development of efficient belief propagation algorithms for template models enabled an explosion of research and applications of probability models in intelligent systems (e.g., Pearl 1988; Jensen, 1996). As belief network technology is applied to more complex problems, the limitations of template models become clear. Even when a domain can be represented by a template model, its size and complexity may make it necessary to represent it implicitly as a collection of modular subunits from which smaller submodels are constructed for reasoning about problem instances (Pradhan et al, 1994). In more complex domains template models are insufficient as a knowledge representation because the relevant variables and their interrelationships vary from problem instance to problem instance. In such domains, belief networks can still be used to capture stable patterns of probabilistic relationships for pieces of the domain, and these pieces brought together to build probability models to reason about particular problem instances (Wellman, Breese and Goldman, 1992; Goldman and Charniak, 1993). There has been a steady interest in automated construction of belief network models in fields such as natural language understanding (Goldman and Charniak, 1993), military situation assessment (Laskey et al, 1993), image understanding (Levitt, et al., 1990), financial securities trading (Breese, 1987), and plan projection (Ngo et al, 1996).

This paper presents a knowledge representation framework to support automated model construction of problem-specific models from a knowledge base expressing generic probabilistic relationships. Most work on automated network construction takes as the unit of knowledge a set of probabilistic influences on a single variable. That is, an element of the knowledge base specifies a variable, some or all of its parents, and information used to construct its local distribution in the constructed model. For a number of reasons, it is useful to have the capability to organize domain knowledge in larger chunks. Domain experts often consider a related set of variables together. The ability to represent conceptually meaningful groupings of variables and their interrelationships facilitates both knowledge elicitation and knowledge base maintenance (Mahoney and Laskey, 1996). Also, larger situation specific models tend to include these conceptually meaningful groupings as submodels. Thus, a model construction algorithm can be made more efficient by searching for and instantiating submodels over sets of related variables.

Our representation therefore takes as its basic unit the network fragment, which consists of a set of related variables together with knowledge about the probabilistic relationships among the variables. We discuss how network fragments can be combined to form larger models for reasoning about a given problem instance. Our focus is on the representation of probabilistic knowledge as network fragments and not on algorithms for constructing models from the knowledge base.

## 2. MILITARY SITUATION ASSESSMENT

The application area for our work is the domain of military situation assessment. We give a brief description of this application area, both to illustrate the complexities of the domain and to provide examples for later reference.

A military intelligence analyst is charged with constructing a description of a military situation: who the actors are, where they are located, what they are doing, and what they are likely to do in the future. To do this, the analyst uses her knowledge of military doctrine and tactics, knowledge about the capabilities and limitations of military forces and equipment, background information about weather and terrain, and reports about the current situation from various sources including radar, imagery, communications traffic, and human informants. Reasoning is performed at different levels of aggregation. For example, an SA6 surface-to-air missile regiment is comprised of several batteries and a command post, and each of these subunits is itself comprised of elements such as launchers, reloaders and radars. For some purposes the analyst may reason about a regiment in the aggregate; for other purposes she may reason about the individual subunits (batteries and command post) comprising the regiment. The analyst must also reason about the evolution of the situation in time.

As an illustration, consider an analyst who has received a report R3 of a radar emission characteristic of a Straight Flush radar. The report is accompanied by an error ellipse which indicates a region within which the radar may be located. A Straight Flush radar is characteristic of a surface-to-air missile battery of type SA6. The analyst considers her current situation model, focusing on the area within the error ellipse of the report. She had previously received an imagery report R2 indicating a unit of unidentified type within the ellipse. The analyst considers the hypothesis that reports R3 and R2 refer to the same unit. In addition, there was a prior report R1 of a straight flush radar. The two error ellipses show little overlap. The analyst therefore considers it possible but unlikely that R1 and R3 came from the same unit at the same location. The report R1 was received several hours ago, so the analyst considers whether the two reports came from a single battery that moved during the time between the reports. Yet another possibility the analyst considers is that the report came from a new, not yet observed, SA6 battery. Under each of these possibilities for the unit giving rise to the report, the analyst must consider the aggregation of the SA6 batteries in the region into regiments. Batteries in a regiment are typically spaced so that there is some overlap in the airspace they are covering, and so that they provide the widest possible area of coverage. She also considers various possibilities for the military target or region the regiment is defending.

This brief vignette covers only a small subset of the reports our analyst receives about the situation over the course of a day. Each report must be considered in the light of her current view of the situation and used to refine her estimate of what is happening. She must reason not just about the current situation but also about how it is likely to evolve. Her description of the situation provides input to her commander, who must plan a course of action to respond to what the opposing force is likely to do.

It is clear that a template model is inadequate for this problem. The number of actors of any given type is not static, but varies from situation to situation. A reasoning system must be capable of unifying reports with already-hypothesized units and/or hypothesizing new units, as the current problem context demands. The relevant variables for reasoning about an actor depend on the type of actor it is. For example, the mode in which a radar emits is a key variable for inferring the activity of a surface-to-air missile battery. However, this variable is simply not applicable to units which have no radar. Clearly a network with a fixed set of variables and a fixed topology is inadequate for this problem.

## 3. NETWORK FRAGMENTS

### 3.1 NETWORK FRAGMENTS AS OBJECTS

We have found it useful to express our representation framework in the language of object-oriented analysis (Rumbaugh, et al., 1991). An advantage of the object-oriented approach is the ability to represent abstract types. Objects of a given type share structure (common attributes) and behavior (common methods). Another important feature is inheritance. Objects can be organized in hierarchies of related objects. From an implementation viewpoint, this facilitates knowledge base development and maintenance. It is much easier to specify a new object type, especially one similar to an existing object type, when much of its structure and behavior are inherited from its parent in the object hierarchy. Maintenance is simplified because changes to structure or behavior need be made only at the level of the hierarchy at which the knowledge is specified, and automatically propagate to all objects inheriting from the changed object. Another advantage of the object-oriented approach is the ability to encapsulate private knowledge within an object. In a related paper, Koller and Pfeffer (1997) discuss the role of encapsulation in the design of large, complex belief network knowledge bases. Finally, objects provide a natural way to represent first-order knowledge about families of problem-specific models. Object classes are used to represent generic knowledge about types of domain entity. In a given problem situation, one or more instances of an object class may be created to reason about particular entities of a given type.

In our framework, there are two basic categories of object: the random variable and the network fragment. Random variables represent the uncertain propositions about which the system reasons; network fragments represent probabilistic relationships between these propositions. Random variable and network fragment classes represent knowledge about generic domain entities. During problem solving, instances of these random variable and network fragment classes are created in a model workspace to represent attributes of particular domain entities and their interrelationships.

## 3.2   RANDOM VARIABLES

Random variables represent aspects of a situation about which the reasoner may be uncertain. Each random variable class has a set of identifying attributes, which are bound to particular values when an instance of the random variable is created. For example, the random variable class (SA6 Battery Activity <*Unit-ID*> <*t*>) represents the activity of an SA6 battery. Its identifying attributes are <*Unit-ID*>, which refers to the particular unit, and <*t*>, which refers to when the activity is taking place. These variables are bound to particular values when an instance is created to refer to a particular situation.

**Definition 1:** A *random variable* is an object with the following attributes and methods:

- *Name*. This is a unique name for the variable class.
- *States*. Our current representation assumes that a random variable has a fixed finite set of possible states. This could be generalized to allow a random variable to have an associated method for determining its state space for the context in which it is instantiated.
- *Identifying attributes*. Each random variable has a set of identifying attributes. These attributes are bound to specific values when the random variable is instantiated.
- *Influence combination*. This is a method for constructing the local distribution of the random variable from probability information contained in multiple fragments. A commonly used example of an influence combination method is the noisy-OR. Influence combination is discussed in more detail in Section 4 below.
- *Default distribution*. This is a method for assigning a distribution to the random variable by default when none is explicitly specified.

As is common with the term object, the term random variable may be used to refer either to a class or an instance. When the intent is not clear from the context, the more specific term random variable class or random variable instance will be used.

## 3.3   ELEMENTARY FRAGMENTS

Network fragments organize sets of random variables and encode the probabilistic relationships among them. The knowledge base designer encodes knowledge in the form of elementary fragment classes, which are instantiated and combined during problem solving into compound fragments. An elementary fragment is a modular, semantically meaningful unit of probability knowledge. Variables within the fragment are classified as *resident* or *input* variables. Distributional information for resident variables is represented within the fragment. Input variables are variables that condition resident variables, but whose distributions are carried external to the fragment.

In our domain it is important to be able to express asymmetric independence, or independence relationships between variables that exist only for certain values of other variables (cf., Geiger and Heckerman, 1991; Boutilier et al., 1996). Our framework generalizes the Bayesian multinet, defining a multi-fragment as a collection of *hypothesis-conditioned fragments* that together specify the distribution of a set of resident variables. Hypothesis-conditioned fragments express knowledge about their input and resident variables conditional on a subset of the state space of hypothesis variables. Hypothesis-conditioned fragments allow parsimonious expression of independence relationships that exist conditional on subsets of the hypothesis variables, but not unconditionally.

A fragment has a set of associated identifying attributes, which map to the identifying attributes of its random variables. For example, Figure 1c is an instance of a hypothesis-conditioned fragment class for reasoning about activity and dwell (length of time at a given location) of a surface-to-air missile unit. The identifying attributes of the fragment correspond to the unit identifier and the current and previous time periods, and are bound to the values <B654>, <0>, and <1> respectively. The unit identifier attribute points to the corresponding attribute in each of the fragment's random variables. This constrains all random variables in the fragment to refer to the same unit.

**Definition 2:** An *elementary hypothesis-conditioned fragment* is an object with the following attributes and methods:

- *Random variables*. Each random variable associated with a fragment takes as its value a random variable instance of a given random

variable class. A nonempty subset R of the fragment's random variables is designated as *resident* variables; the remaining (possibly empty) set I of random variables is designated as *input* variables. A subset H of the input variables of fragment F is designated as the *hypothesis variable set*.

- *Hypothesized* subset. A subset μ of the Cartesian product of the state spaces of the hypothesis variables is designated as the *hypothesized subset* for fragment F.

- A set of *fragment identifying attributes* and a mapping from the fragment identifying attributes to the identifying attributes of the fragment random variables. These identifying attributes play the role of variables in a logic programming language.

- An acyclic directed graph G over I×R called the *fragment graph*, in which all nodes in I are root nodes;

- An *influence function* for each variable in the fragment. The influence function is used by the influence combination method to compute a local distribution for the variable.

- A *local distribution* for each resident variable in the fragment. The local distribution represents a probability distribution over the state space of the variable given each combination of values of its parents in G.

The local distribution need not be represented explicitly. When a fragment represents a partial influence and contains only a subset of the parents of the variable, the local distribution attribute may be left unspecified or may contain a default distribution to be used when only the parents mentioned in the fragment are included in the constructed model. A model construction system need not compute local distributions until it is ready to use the model for inference.

## 4 FRAGMENT COMBINATION

### 4.1 INTRODUCTION

This section describes the process of combining fragment instances into larger models for reasoning about a problem. Figure 1 shows an example of fragment combination for fragments used in reasoning about where an SA6 battery is located and how long it will remain at that location. Figures 1a and 1b focus on location quality. Location quality is important for inference about location because units tend to be placed where location quality is good. The fragment instances in Figure 1a and Figure 1b represent the partial influence on location quality of the degree to which a location supports the unit's mission and the degree to which the location supports its activity. Both these influences are mediated by the unit's activity. These influences are combined with a conditional noisy-MIN influence combination, in which the influences of the two supportability variables combine by a leaky noisy-MIN for each value of the activity variable. Figure 1c is an instance of a fragment expressing knowledge about the unit's activity and how long the unit will remain at its present location. These fragments are combined into the result fragment shown in Figure 1d.

### 4.2 INFLUENCE COMBINATION

When fragments are combined, local distributions for the combined fragment are computed from the fragment influence functions using the node's influence combination method. The influence function for a variable in a fragment in which it is resident must provide the inputs needed by that variable's influence combination method. Thus, influence combination and influence functions must be designed to work together. A number of generic influence combination methods have appeared in the literature. We describe several common methods below.

The most straightforward influence combination method is Simple-Combination, which requires the variable X to be resident in exactly one fragment containing all its parents. The influence function for X computes its (possibly unnormalized) local distribution, and Simple-Combination simply normalizes this distribution. Using Simple-Combination, it is straightforward to represent a standard Bayesian network over $n$ variables $X_1, ..., X_n$ as a set of $n$ network fragments. Each fragment $F_i$ has exactly one resident variable, $X_i$. The input variables of $F_i$ are the parents of $X_i$ in the original Bayesian network. These fragments combine to yield the original Bayesian network. Slightly more complex than Simple-Combination is Default-Combination, in which a default distribution is overridden by a distribution defined for a more specific set of parent variables.

Another class of influence combination methods consists of methods for combining partial influences. The most commonly cited partial influence models are the independence of causal influence (ICI) models[1], the best-known of which is the noisy-OR. For an ICI model, the influence function carries information about the partial influence of a subset of the node's parents. When several fragments expressing partial influences are combined, the node's influence combination method uses the partial influence information computed by each fragment's influence function to compute a local distribution given all the parents. The fragments of Figure 1a and 1b are combined using a modified ICI method.

Another generic type of influence combination, Parameterized-Combination, occurs when, again, X is resident in a single fragment containing all its parents, but its distribution can be computed from some lower dimensional representation. One such example is the sigmoid function (Jaakkola and Jordan, 1996; Neal, 1992). When the set of influences is known in advance, partial influence models may also be represented using a single home fragment and Parameterized-Combination. The Parameterized-Combination influence function returns the parameters used to compute the local distributions, and influence combination computes the local distribution from the parameters.

---

[1] Independence of causal influence has also been called causal independence (see Heckerman, 1993).

An influence combination method has a set of enabling conditions specifying requirements for applicability of the method. The enabling conditions provide a way for the designer to specify conditions under which the combination method applies. For example, all input nodes to a noisy-OR must be binary, as must the node for which the distribution is being computed. As another example, the method Simple-Combination completes without error only when the variable is resident in exactly one fragment of the input set. Assuming that its enabling conditions are met, the influence combination method computes the variable's local distribution using the results returned by the variable's influence functions from the input fragments in which it is resident.

Combining hypothesis-conditioned fragments requires conditions involving consistency of their hypothesized subsets. Hypothesis-conditioned fragments are organized into multi-fragments (Section 5), which consist of a partition over a set of hypothesis variables together with a set of hypothesis-conditioned fragments defining distributions for resident variables given the hypothesis variables. For this reason, an influence combination method for a variable X takes as inputs not only the fragments whose distributions for X are to be combined, but also the partition element for which the ouptut distribution is being computed. The following definitions establish terminology for the consistency conditions influence combination is required to satisfy.

***Definition 3:*** An *hypothesis partition* $S=(H, \Delta)$ is a set of variables H together with a partition $\Delta$ of the Cartesian product of the state spaces of variables in H. An *hypothesis element* of the hypothesis partition is an element $\nu$ of $\Delta$.

***Definition 4:*** Let F be an hypothesis-conditioned fragment instance with resident variables $R_F$, input variables $I_F$, hypothesis variables $H_F$ and hypothesized subset $\mu_F$. Let $S=(H, \Delta)$ be a hypothesis partition and let $\nu \in \Delta$ be a hypothesis element. The fragment F and the hypothesis element $\nu$ are *hypothesis variable consistent* if: (1) $H_F \subset H$ and (2) if $X \in H$ and $X \in (I_F \cup R_F)$ then $X \in H_F$. F *subsumes* $\nu$ if $\nu \subset \mu_F$. F is *disjoint from* $\nu$ if $\nu \cap \mu_F = \emptyset$.

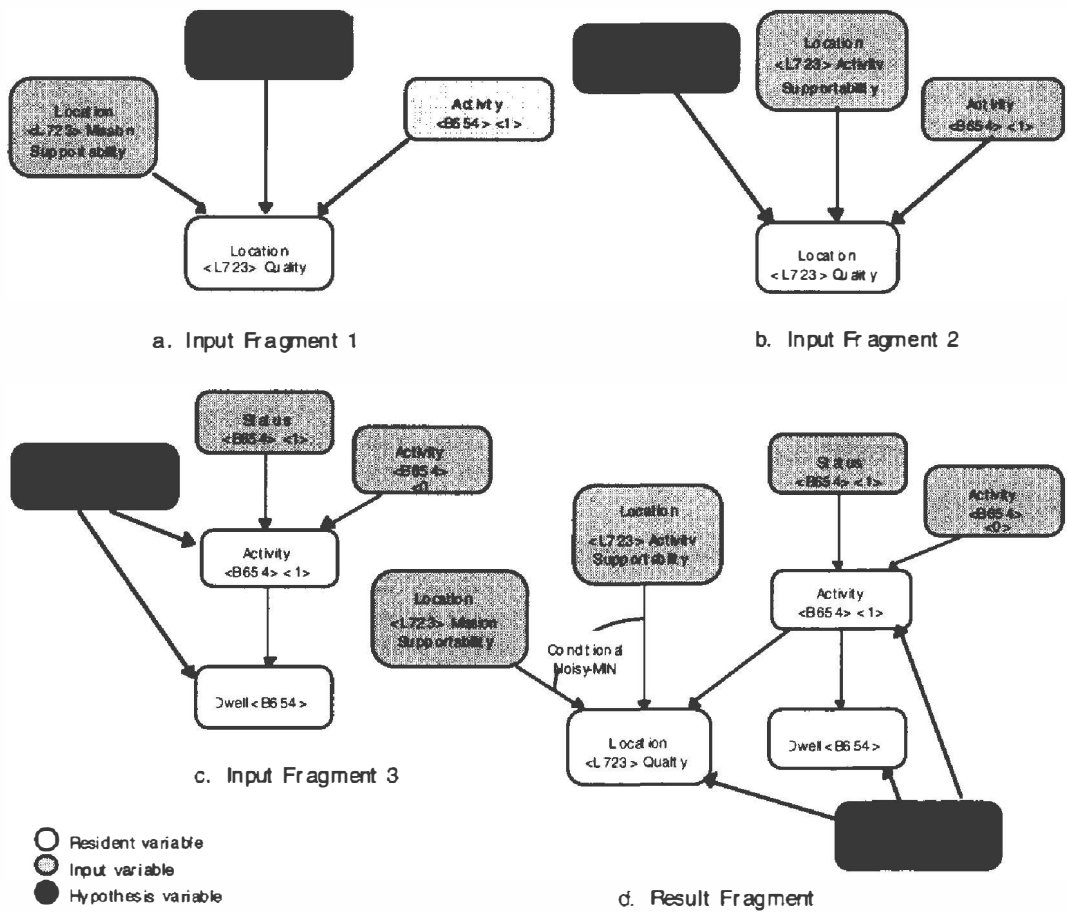

Figure 1: Example of Fragment Combination

Hypothesis variable consistency simply means that the fragment and the hypothesis partition agree on which variables are designated as hypothesis variables. All variables in the hypothesis partition that appear in the fragment must be designated there as hypothesis variables. Moreover, any variable designated in the fragment as a hypothesis variable must be included in H. The hypothesis partition of a multi-fragment is required to satisfy hypothesis variable consistency with each of its component hypothesis-specific fragments.

When a fragment subsumes v its hypothesized subset contains v, which implies that the fragment defines local distributions for its resident variables given each state of v. Each resident variable of a multi-fragment must be resident in a hypothesis-conditioned fragment subsuming v for each hypothesis element v of the multi-fragment's hypothesis partition. This condition ensures that complete local distributions are specified for all resident variables in the multi-fragment.

Finally, for each hypothesis-conditioned fragment F and each hypothesis element v, F must either subsume v or be disjoint from v. This condition ensures that if a distribution is defined by F for some states in v, then F defines distributions for all states in v. If this condition is not satisfied by a set D of fragments and an hypothesis partition $S=(H,\Delta)$, there exists a refinement $\Delta'$ of $\Delta$ for which it is satisfied.

***Definition 5:*** Let X be a node and let $S = (H, \Delta)$ be an hypothesis partition. An *influence combination method* for X is a function which takes as input a set D of hypothesis-conditioned fragments and an hypothesis element $v \in \Delta$, and which satisfies:

- An error is returned unless: (1) X is resident in at least one fragment in D subsuming v; (2) X is resident only in fragments in D which either subsume v or are disjoint from v; and (3) the enabling conditions specific to the influence combination method are satisfied.

- Otherwise, the function returns a set of parents for X and a local distribution for X.

- The parents returned for X are the variables containing arcs into X in the graph union of the fragment graphs for fragment instances in D subsuming v.

- The local distribution for X is computed using the influence functions for X from the fragment instances in which X is resident and that subsume v.

- The parents and local distribution returned for X depend only on those fragments in D in which X is resident and which subsume v.

The following definitions provide conditions under which a set of fragments can be combined into a compound fragment that unambiguously defines a probability distribution over its resident variables given its inputs.

***Definition 6:*** Let $S=(H, \Delta)$ be an hypothesis partition, $v \in \Delta$ an hypothesis element, and D a set of hypothesis-conditioned fragment instances. D is *acyclic* given v if the graph union of the fragment graphs for all fragment instances in D that subsume v contains no directed cycles.

***Definition 7:*** Let X be a random variable instance, $S=(H, \Delta)$ an hypothesis partition, $v \in \Delta$ an hypothesis element, and D a set of hypothesis-conditioned fragment instances. D satisfies *home fragment consistency* for X and v if the influence combination method for X returns without error for D and v.

***Definition 8:*** Let $S=(H, \Delta)$ be an hypothesis partition, $v \in \Delta$ an hypothesis element, and D a set of hypothesis-conditioned fragment instances. Then D is *globally consistent* given v if the following conditions are satisfied for each X resident in at least one fragment of D that subsumes v: (1) D is acyclic given v; (2) F and v have consistent hypothesis partitions for each $F \in D$; (2) D is home fragment consistent for X and v.

### 4.3 COMPOUND FRAGMENTS

A globally consistent set of fragment instances can be combined into a compound fragment as defined below. Compound fragments differ from elementary fragments in that compound fragments have no influence functions of their own, but point to their component fragments where the influence functions reside. The local distribution for a variable in the compound fragment is computed by calling the variable's influence combination method, which in turn calls the variable's influence functions from the component fragments in which the variable is resident. Maintaining pointers to the component fragments facilitates incremental model construction and permits computation of the local distributions to be deferred until needed by the inference algorithm.

***Definition 9:*** Let $S = (H, \Delta)$ be a hypothesis partition, let $v \in \Delta$ be a hypothesis element, and let D be a globally consistent set of hypothesis-conditioned fragment instances. The *compound fragment* $F_{D,v}$ is an object with the following attributes:

- *Hypothesis variables* H;
- *Hypothesis element* v;
- *Resident variables* R consisting of those variables resident in at least one fragment in D;
- *Input variables* I consisting of variables that are input to at least one fragment in D and resident in no fragment in D.
- *Fragment graph* consisting of an acyclic directed graph defined as follows. The nodes in G are given by $I \cup R$, where I and R are defined above. All nodes in I are root nodes. The parents of a node X in R are the variables returned as parents by the influence combination method of X applied to D.

- *Component fragments* consisting of all elementary fragments in D together with all component fragments of compound fragments in D.
- *Local distribution* for resident variables. These may be left unspecified. If specified, the local distribution for X is computed by applying the influence combination method for X to D.

It is clear from the above definition that fragment combination is order-independent. It may be useful when models are constructed incrementally to permit the knowledge base designer to define incremental influence combination methods. Incremental influence combination would compute the local distribution for a compound fragment from the local distributions of input compound fragments, together with the influence functions of input elementary fragments.

## 5. MULTI-FRAGMENTS

Representing knowledge as hypothesis-conditioned fragments is convenient when a different fragment graph structure applies for different states of the hypothesis variables. To represent such a model as a standard Bayesian network or network fragment would require a more complex structure than the individual, simpler structures associated with the subsets. For some problems, knowledge representation, knowledge elicitation, and data entry may be significantly simplified by the hypothesis-conditioned fragment representation. Most of the models in our current knowledge base are hypothesis-conditioned fragments, and many of the interesting inference tasks require combining these hypothesis-conditioned fragments into multi-fragments. For example, the imagery report R2 described in Section 2 refers to a unit of unknown type. One possibility for the unit's type is an SA6 battery. The hypothesis-conditioned fragments of Figure 1 would be retrieved for reasoning about the unit's activity and location under the hypothesis that it is an SA6 battery as well as fragments for the other possibilities for the unit's type.

As in a Bayesian multinet, all resident variables in a multi-fragment must have distributions defined for all hypotheses in the multi-fragment's hypothesis partition. In our domain, there are many variables that exist only for some values of a hypothesis (e.g., radar mode, which is only defined if the unit is a type which has a radar). We handle these variables by defining their state as the special state NA in hypothesis-conditioned fragments in which the variable is not defined.

Hypothesis-conditioned fragments may combined by multi-fragment combination as defined below.

*Definition 10:* Let S=(H, $\Delta$) be a hypothesis partition and let D be a set of hypothesis-conditioned fragment instances that is globally consistent given v for each hypothesis element v$\in \Delta$. Then the *multi-fragment* with component fragments D and hypothesis partition is an object with attributes:

- *Resident variables:* the set R of variables resident in at least one fragment in D.
- *Input variables:* the set I of variables that are input to at least one fragment in D and resident in no fragment in D.
- *Fragment graph*: an acyclic directed graph defined as follows. The nodes in G are given by I$\cup$R, where I and R are defined above. All nodes in I are root nodes. The parents of a node X in R are all variables returned as parents by the influence combination method of X applied to D and v for some hypothesis element v.
- *Component fragments:* all elementary hypothesis-conditioned fragment instances in D together with all component fragments of compound hypothesis-conditioned fragment instances in D.
- *Local distributions:* for each resident variable X a local distribution may be represented explicitly with the multi-fragment. If specified, it is computed by applying the influence combination method for X to D and v for all hypothesis elements v.

A multi-fragment defines a probability distribution over its resident variables given its input variables. The multi-fragment representation permits a knowledge base designer to exploit asymmetric independencies in a domain to specify a set of interrelated, structurally simple submodels that together comprise a probability model for a domain. Generally, the variables appearing as resident variables in a given multi-fragment will be ones for which the given partition of the hypothesis variables induces a simple network structure on the constituent fragments. Sometimes different partitions will induce simple structures for different sets of child variables. When this is the case, different multi-fragments may be defined over these different sets of variables. Multi-fragments may be combined with other multi-fragments to form compound fragments in a straightforward extension to Definition 9.

## 6. MODEL CONSTRUCTION

Model construction proceeds by retrieving fragment classes from a knowledge base, creating fragment instances in the model workspace, and combining them by the operations defined above. A model in the model workspace represents a complete probability model over its variables when the set of input variables is empty. A model is *query complete* for query Q = $P(X_t|X_e=x_e)$ if the evidence variables $x_e$ d-separate the target variables $X_t$ from the input variables. The provision for default distributions for input variables permits approximate reasoning using incomplete models, as needed for anytime model construction.

For knowledge bases encoding modularized template models, model construction means selecting which parts of the template model to bring into the model workspace. Variables that are $d$-separated by observed variables from target variables need not be explicitly represented. Some search algorithms involve computing or approximating bounds on the influence of a variable to decide whether the computation involved in extending the model is justified by the potential improvement in accuracy.

In our application, model construction involves additional issues, among them the problems of data association, hypothesis management, and pattern replication. Data association is the problem of deciding which domain entity a piece of evidence refers to. An example of data association is reasoning about which already-hypothesized SAM unit, if any, should be associated with an intelligence report indicating a SAM unit. Hypothesis management is the problem of generating and pruning hypotheses about domain entities and their interrelationships. Pattern replication refers to the need to make multiple copies of a model to refer to different domain entities or different instants in time for a temporal reasoning problem. Our representation framework was developed to support these model construction functions, although they are not treated in the present paper.

We have implemented a simplified version of the fragment combination operations of Section 4 in the PRIDE® system, developed a library of fragments for the situation assessment domain, and are developing an object-oriented database schema for our fragment library.

## Acknowledgments

The research reported in this paper was sponsored by DARPA and the U.S. Army Topographic Engineering Center under contract DACA76-93-0025 to Information Extraction and Transport, Inc. The authors extend grateful acknowledgment to Tod Levitt, Daphne Koller, and three anonymous reviewers for helpful comments and suggestions on earlier versions of this paper.